\documentclass[11pt,a4paper]{article}
\usepackage[hyperref]{eacl2021}
\usepackage{times}
\usepackage{latexsym}

\usepackage{url}
\usepackage{microtype}
\usepackage[utf8x]{inputenc}
\usepackage{amsthm}
\usepackage{adjustbox}
\usepackage{multirow}
\usepackage{multicol}
\usepackage{arydshln}
\usepackage{graphicx}
\usepackage{subcaption}
\usepackage{mathtools}
\usepackage{amsmath}
\usepackage{amssymb}
\usepackage{enumerate}
\usepackage{booktabs}

\aclfinalcopy \def\aclpaperid{065}

\title{Effects of Pre- and Post-Processing on type-based Embeddings in \\Lexical Semantic Change Detection}

\author{Jens Kaiser\thanks{~Authors contributed equally, and their ordering was determined randomly.}, Sinan Kurtyigit\footnotemark[1], Serge Kotchourko\footnotemark[1], Dominik Schlechtweg\\
Institute for Natural Language Processing, University of Stuttgart\\
\small
{\tt \{jens.kaiser,sinan.kurtyigit,serge.kotchourko,schlecdk\}@ims.uni-stuttgart.de}}

\date{}

\begin{document}
\maketitle
\begin{abstract}
Lexical semantic change detection is a new and innovative research field. The optimal fine-tuning of models including pre- and post-processing is largely unclear. We optimize existing models by (i) pre-training on large corpora and refining on diachronic target corpora tackling the notorious small data problem, and (ii) applying post-processing transformations that have been shown to improve performance on synchronic tasks. Our results provide a guide for the application and optimization of lexical semantic change detection models across various learning scenarios.
\end{abstract}

\section{Introduction}
In recent years Lexical Semantic Change Detection (LSCD), i.e. the detection of word meaning change over time, has seen considerable developments \citep{2018arXiv181106278T,kutuzov-etal-2018-diachronic,hengchen2021challenges}. The recent publication of multi-lingual human-annotated evaluation data from SemEval-2020 Task 1 \citep{schlechtweg2020semeval} makes it now possible to compare LSCD models in a variety of scenarios. 
The task shows a clear dominance of type-based embeddings, although these are strongly influenced by the size of training corpora. In order to mitigate this problem we propose pre-training models on large corpora and refine them on diachronic target corpora.
We further improve the obtained embeddings with several post-processing transformations which have been shown to have positive effects on performance in semantic similarity and analogy tasks \citep{mu-etal-2017-representing,artetxe-etal-2018-uncovering,raunak-etal-2019-effective} as well as term extraction \citep{Haettyetal20}. Extensive experiments are performed on the German and English LSCD datasets from SemEval-2020 Task 1. According to our findings, pre-training is advisable when the target corpora are small and should be done using diachronic data. We further show that pre-training on large corpora strongly interacts with vector dimensionality and propose a simple solution to avoid drastic performance drops. Post-processing often yields further improvements. However, it is hard to find a reliable parameter that performs well across the board. Our experiments suggest that it is possible to use simple pre- and post-processing techniques to improve the state-of-the-art in LSCD.

\section{Related Work}
As evident in \citet{schlechtweg2020semeval} the field of LSCD is currently dominated by Vector Space Models (VSMs), which can be divided into type-based (static) \citep{Turney:2010} and token-based (contextualized) \citep{Schutze1998} models.
Prominent type-based models include low-dimensional embeddings such as Global Vectors \citep[GloVe,][]{pennington-etal-2014-glove} and Skip-Gram with Negative Sampling \citep[SGNS,][]{Mikolov13a, Mikolov13b}.
However, as these models come with the deficiency that they aggregate all senses of a word into a single representation, token-based embeddings have been proposed \citep{peters-etal-2018-deep,devlin-etal-2019-bert}.
According to \citet{Hu19} these models can ideally capture complex characteristics of word use, and how they vary across linguistic contexts. The results of SemEval-2020 Task 1 \citep{schlechtweg2020semeval}, however, show that contrary to this, the token-based embedding models \citep{beck-2020-diasense, kutuzov-giulianelli-2020-uiouva} are heavily outperformed by the type-based ones \citep{prazak-etal-2020-uwb, asgari-etal-2020-emblexchange}. The SGNS model was not only widely used, but also performed best among the participants in the task. This result was recently reproduced in the DIACR-Ita shared task \citep{diacrita_evalita2020, laicherevalita2020, kaiserdiacrita2020}. Its fast implementation and combination possibilities with different alignment types further solidify SGNS as the standard in LSCD \citep{schlechtweg2020semeval, Schlechtwegetal19, Shoemark2019, kutuzov2020ShiftRy}. Hence, the embeddings used in this work are SGNS-based. 

Further increases in performance of type-based VSMs can be achieved by various post-processing transformations. This has been shown for semantic similarity and analogy tasks \citep{mu-etal-2017-representing,artetxe-etal-2018-uncovering,raunak-etal-2019-effective} as well as term extraction \citep{Haettyetal20}. It is still an open question whether these transformations improve performance in the special setting of LSCD where we typically have several corpora and vector spaces which have to be transformed simultaneously \citep{schlechtweg2020semeval}. 
An indication is given by \citet{Schlechtwegetal19} showing that for a simple LSCD model mean centering leads to consistent performance improvements on two German data sets. Whether this result is reproducible on further data sets, more complex models and further post-processing techniques has not been determined yet.

Post-processing methods operate on information already contained in a VSM, rather than adding additional information. Further semantic information can be introduced by pre-training vectors on a larger unspecific collection of text \citep{kutuzov2016cross} or by training a seperate matrix on such text and concatenating the two VSMs \citep{limsopatham-collier-2016-modelling}. This is especially helpful for cases where only smaller specialized corpora are given. Combining the information from two models is also found in \citet{Kim14}, here it is used for alignment proposes. We operate similarly to \citeauthor{Kim14} but with the motivation of \citeauthor{limsopatham-collier-2016-modelling} and \citeauthor{kutuzov2016cross}, as we aim to enrich a VSM prior to the training process.

\section{Data and Tasks}

\begin{table*}[ht]
\center
\small
\begin{tabular}{l|cccc|cccc}
\toprule
 & \multicolumn{4}{c}{\textsc{diachron}} & \multicolumn{3}{|c}{\textsc{modern}} \\
& GER$_{t1}$ & GER$_{t2}$ & ENG$_{t1}$ & ENG$_{t2}$ &  \textsc{SdeWaC} & \textsc{PukWaC}  \\
\hline
source & \multicolumn{1}{c}{DTA}  & \multicolumn{1}{c}{BZ+ND} & \multicolumn{1}{c}{CCOHA}  &  \multicolumn{1}{c |}{CCOHA}  & \multicolumn{1}{c}{web} &  \multicolumn{1}{c}{web} \\
\multirow{2}{*}{time period} & 1800 -- & 1946 -- & 1810 -- & 1960 -- & $\sim$2005 -- & $\sim$2005 -- \\ [-0.4em]
& \quad 1899 & \quad 1990 & \quad1860 & \quad2010 & \quad 2005 & \quad 2005 \\ [+0.4em]
\# of tokens & 66.9M & 67.2M & 6.48M & 6.62M & 750M & 1.92B \\
\# of types & 51.1K & 59.1K & 25.9K & 37.5K & 44.6K & 51.9K \\ 
min word freq. & 39 & 39 & 4 & 4 & 450 & 750 \\
\bottomrule
\end{tabular}
\caption{Corpus statistics. GER$_{t1}$ and GER$_{t2}$ are sampled from DTA \cite{dta2017}, BZ \cite{BZ2018} and ND \cite{ND2018}. DTA contains texts from different genres, BZ and ND are collections of newspaper articles. Clean Corpus of Historical American English (CCOHA) \citep{Davies:2012, Alatrashetal20} is a genre balanced collection of texts from a wide variety of time periods and the basis for ENG$_{t1}$ and ENG$_{t2}$.}
\label{tab:corpora}
\end{table*}

\label{sec:data}
We train SGNS-based VSMs on various corpora and use a word similarity task and an LSCD task for evaluation. 
The two tasks share a common aspect: the vector representations of two words need to be compared with some metric (e.g. cosine similarity), and word pairs need to be ranked according to that metric. In the word similarity task, we have the vectors of two different words in the same vector space $(w_i, w_j)$, while for LSCD we have the vectors of the same word but from different vector spaces representing different time periods $(w_i^{t1}, w_i^{t2})$.

\paragraph{Modern Data.} We use two large modern English and German corpora, \textsc{PukWaC} \citep{Baroni2009} and \textsc{SdeWaC} \citep{faasseckart2012} to validate the post-processing methods on the word similarity task and to create pre-trained embeddings for the LSCD task. \textsc{PukWaC} and \textsc{SdeWaC} are web-crawled corpora from the \textit{.uk} and \textit{.de} domain respectively. Resulting in fairly large corpora, ~2B tokens and ~750M tokens (see Table \ref{tab:corpora}). We evaluate vector representations created on the two corpora on a standard dataset of human similarity judgments, WordSim353 \citep{Finkelstein2002}, by measuring Spearman's rank correlation coefficient of the cosine similarity of vectors for target word pairs with human judgments.

\paragraph{Diachronic Data.} We utilize the English and German datasets provided by SemEval-2020 Task 1 Subtask 2 \citep{schlechtweg2020semeval}. Each dataset contains two \textit{target corpora} from different time periods, $t_1$ and $t_2$, as well as a list of target words. The corpora originate mostly from newspaper articles and books. Their biggest difference to \textsc{PukWaC} and \textsc{SdeWaC} is their approximately 10 to 100 times smaller size, according to token counts (see to Table \ref{tab:corpora}). The task is to rank the list of target words according to their word sense divergence, gradually from 0 (no change) to 1 (total change). The rank predictions are compared against gold data which is based on human judgments. Once again Spearman's rank correlation coefficient is used to measure performance on the task.

\section{Models}
Following the popular approach taken for type-based vector space models in LSCD, we combine three sub-systems: (i) creating semantic word representations, (ii) aligning them across corpora, and (iii) measuring differences between the aligned representations \citep{Schlechtwegetal19, dubossarskyetal19, Shoemark2019}. Alignment is needed as columns from different vector spaces may not correspond to the same coordinate axes, due to the stochastic nature of many low-dimensional word representations \citep{Hamilton2016b}. Additionally, we aim to refine sub-system (i) by adding pre-trained semantic word representations and using post-processing methods to improve the quality of the created semantic word representations.\footnote{Find a comprehensive overview of type-based LSCD models including semantic representations, alignments and measures in \citet{Schlechtwegetal19}.}

We use SGNS \citep[][]{Mikolov13a, Mikolov13b} to create type-based word representations in combination with three different alignment methods, Orthogonal Procrustes (OP), Vector initialization (VI), and Word Injection (WI). The three alignment methods combined with SGNS have been proven to be state-of-the-art, even when competing against token-based embeddings \citep{schlechtweg2020semeval, kaiser-etal-2020-IMS, diacrita_evalita2020}. Cosine Distance (CD) is used to measure differences between word vectors.\footnote{We provide our code at: \url{https://github.com/Garrafao/LSCDetection}.}

\subsection{Alignment}
\label{sec:alignment}
\paragraph{Vector initialization (VI).} In VI we first train SGNS on one corpus and then use the learned word and context vectors to initialize the model for training on the second corpus \citep{Kim14, kaiser-etal-2020-IMS}. The motivation is that the vector of a word with similar contexts across both corpora will not deviate much from its initialized value. On the other hand, vectors of words with different contexts across both corpora, will be updated to accommodate the new semantic properties. Words which only appear in the second corpus are initialized on random vectors.

\paragraph{Orthogonal Procrustes (OP).} SGNS is trained on each corpus separately, resulting in word matrices $A$ and $B$. To align them, we follow \citet{Hamilton2016b} and calculate an orthogonally-constrained matrix $W^*$: 
\begin{equation*}
W^* =\underset{W \in O(d)}{\arg\min} \left\lVert B W - A\right\lVert_F .
\end{equation*}
Prior to this alignment step both matrices are length-normalized and mean-centered \citep{Artetxe2017,Schlechtwegetal19}.

\paragraph{Word Injection (WI).} The sentences of both corpora are shuffled into one joint corpus, but all occurrences of target words are substituted by the target word concatenated with a tag indicating the corpus it originated from \citep{ferrari2017detecting,Schlechtwegetal19,dubossarskyetal19}. This leads to the creation of two vectors for each target word in one vector space, while non-target words receive only one vector encoding information from both corpora.

\paragraph{No Alignment (NO).} Comparing two vector spaces without aligning them results in poor performance on LSCD \citep{Schlechtwegetal19}. As VI shows, initializing the model with weights from the previous run, results in aligned vector spaces. We expand on this concept by initializing two models on the same pre-trained weights assuming that the resulting vector spaces are aligned to one another. The difference to VI is that instead of initializing model $B$ with the weights from model $A$, the weights from a third pre-trained model $C$ are used to initialize both models $A$ and $B$.

\begin{figure*}[ht]
    \begin{subfigure}{0.5\textwidth}
        \includegraphics[width=\linewidth]{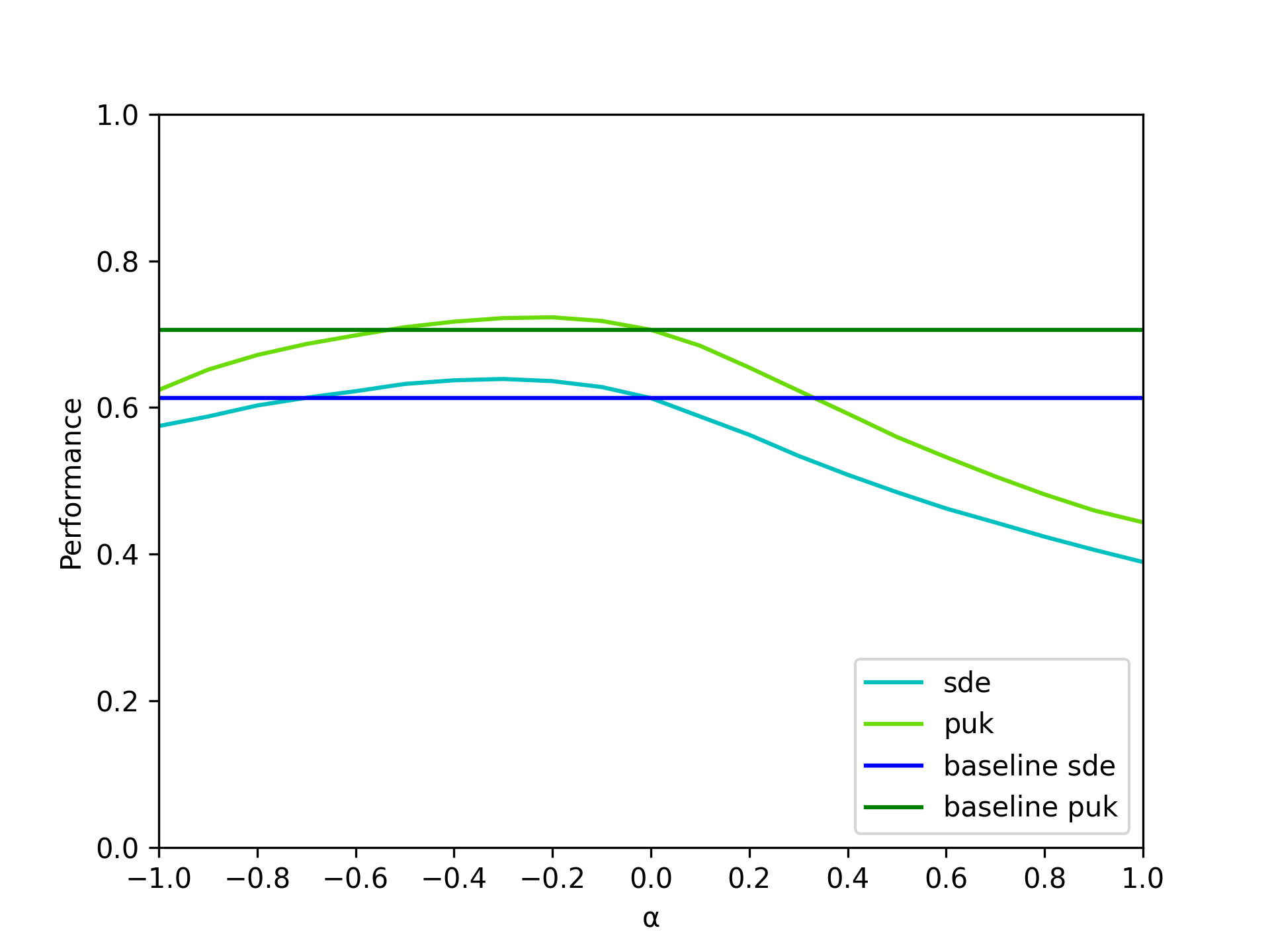}
        \caption{ }
        \label{fig:sot_sdepuk}
    \end{subfigure}
    \begin{subfigure}{0.5\textwidth}
        \includegraphics[width=\linewidth]{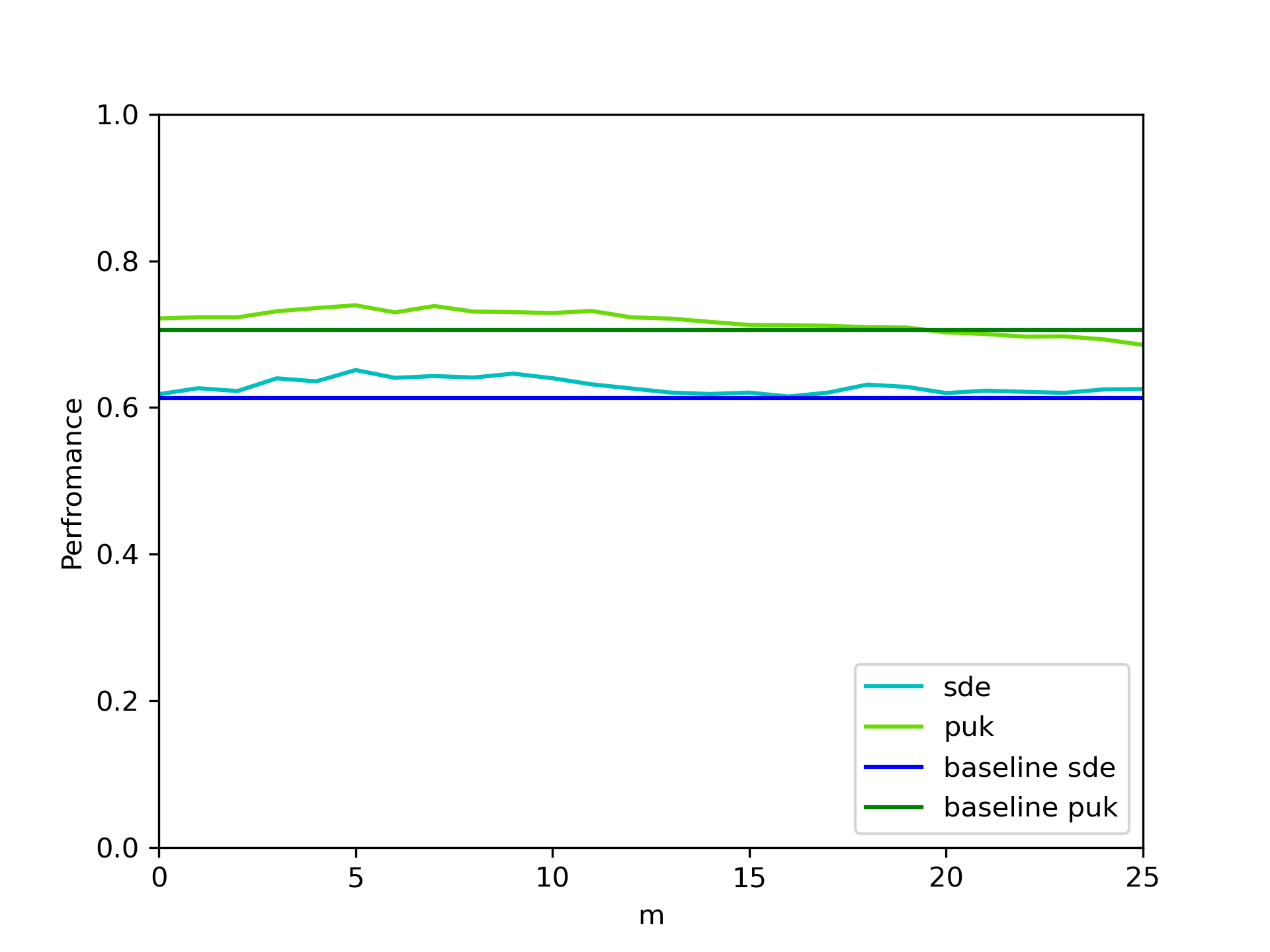}
        \caption{ }
        \label{fig:moc_sdepuk}
    \end{subfigure}
    \caption{Performance on modern data (wordsim353) \textbf{left}: SOT for $\alpha \in [-1,1]$, \textbf{right:} MC+PCR across different amounts of PCs removed. Zero PCs removed indicates only mean centering. Baselines are performances without PP.}
    \label{fig:sde_puk}
\end{figure*}

\subsection{Pre-training}
\label{sec:pretraindescription}
The corpora used in the context of LSCD are often small, as they are restricted by the length of time periods or availability of historical data. For example the English corpora of SemEval-2020 Task 1 only have $6.6$M tokens each, compared to $1.9$G of \textsc{PukWaC}. 
This reduced corpus size limits the amount of semantic information encoded into VSMs trained on the corpus. Pre-training addresses this problem by first training SGNS on a large, possibly external corpus, and then using these vectors to initialize the model for training on the smaller diachronic target corpora. The idea is that the model first learns very broad and general semantic properties followed by the training on the target corpora, where corpus and time specific details are picked up, i.e., a form of refinement. This procedure is applicable to all alignment types.

We use \textsc{PukWaC} and \textsc{SdeWaC} for pre-training, later referenced as \textsc{modern}. However, pre-training on modern corpora is only advisable if the assumption can be made that the meanings of words in the pre-training corpus roughly correspond to the meanings of words in the target corpora. It is unclear to which extent this assumption holds for our data. Hence, we also combine the two target corpora into a bigger corpus, referenced as \textsc{diachron}, which is then used for pre-training.

\subsection{Post-processing (PP)}

\paragraph{Similarity Order Transformation (SOT).}
In 2nd order similarity, the similarity of two words is assessed in terms of how similar they are to a third word \citep{hinjan93, artetxe-etal-2018-uncovering, Schlechtwegetal19SecondOrder}. This can analogously be done for higher (3rd, 4th, etc.) orders. According to \citet{artetxe-etal-2018-uncovering} these orders capture different aspects of language. \citeauthor{artetxe-etal-2018-uncovering} propose a linear transformation deriving higher or lower orders of similarity from a given matrix $X$. For this, the product with the transpose matrix is split into its eigendecomposition $X^TX = Q\lambda Q^T$, so that $\lambda$ is a positive diagonal matrix whose entries are the eigenvalues of $X^TX$ and $Q$ is an orthogonal matrix with their respective eigenvectors as columns. The linear transformation matrix is then defined as $W_\alpha = Q\lambda^\alpha$, where $\alpha$ is the parameter that adjusts the desired similarity order. Applying this to the original embeddings $X$ yields the transformed embeddings $X' = XW^\alpha$.

\paragraph{Mean Centering (MC).}
The centroid of a matrix is the average vector over all vectors in a matrix: $\vec{\bar{c}} = \frac{1}{|V|}\sum_i^V \vec{w_i}$. MC refers to subtracting $\vec{\bar{c}}$ from each $\vec{w_i}$ in the matrix. MC alters all dimensions so that the mean of all columns is zero. \citeauthor{artetxe-etal-2016-learning} provide the intuitive motivation for MC that it moves randomly similar vectors further apart and \citet{mu2018allbutthetop} consider mean centering as an operation making vectors ``more isotropic", i.e., more uniformly distributed across the vector space. \citeauthor{mu2018allbutthetop} indicate that isotropy of word vectors is positively correlated to performance.

\paragraph{Principal Component Removal (PCR).}
Given a $n$-dimensional matrix $X$, Principal Component Analysis \citep[PCA,][]{pearson1901lines} returns $n$ vectors where each vector describes a best fitting line for the data while being orthogonal to the first $n-1$ vectors. Thus, the first PC describes the greatest variance in the first direction, the second PC describes the second greatest variance in the second direction, and the $n$th PC describes the $n$th greatest variance in the $n$th direction. 
\citet{mu2018allbutthetop} use PCA to compute the top $m$ PCs from a mean centered word embedding $\bar{M}$: $p_1, ..., p_m = \text{PCA}(\bar{M})$. Subsequently these PCs are used to project each vector $v \in M$ onto the subspace spanned by the PCs. This projection is then subtracted from the original mean centered word vector $\tilde{v}$ by $v' = \tilde{v} - \sum_{i=1}^m(p_i^\intercal v) p_i$, which results in nullifying the top $m$ PCs in $M$. This is similar to the approach of \citet{bullinariaPCA}. \citeauthor{mu2018allbutthetop} combine both MC and PCR into one PP transformation (MC+PCR).

As for MC \citeauthor{mu2018allbutthetop}'s main motivation for PCR is to make vectors more isotropic. They also demonstrate empirically that the top PCs encode word frequency and offer the removal of this noise from the matrix as an alternative explanation for observed performance improvements.

\paragraph{Stacking.} \label{par_stacking} VI and OP alignment result in two matrices, and hence, a proper way for applying PP to both of them is needed. The naïve way of simply post-processing both matrices separately (SEP) may violate the assumption that they are represented in the same space. Therefore, in a second approach, we apply PP to both matrices simultaneously by stacking them vertically beforehand (STA). Preliminary experiments showed that following the naïve way of PP (SEP) led to severe decrease in performance for SOT (but not for MC+PCR). Hence, applying SOT on two matrices separately is followed by an orthogonal post-alignment (SEP+PA).

\section{Experiments}
\label{sec:experiments}

For the most part, we chose common model hyper-parameter settings in order to keep our results comparable to previous research \citep{Hamilton2016b,Schlechtwegetal19, kaiser-etal-2020-IMS}. We fine-tune for different alignment methods and datasets by varying dimensionality $d$, window size $w$ and number of training epochs $e$.\footnote{For a detailed overview on SGNS parameters see Appendix \ref{sec:parameters}.}

\subsection{Validation}

We validate the results reported by \citet{Artetxe2018a} and \citet{mu2018allbutthetop} on \textsc{PukWaC} and \textsc{SdeWaC}. The performance peaks for negative $\alpha$-values around -0.2 as well as the slight performance increase over the baseline for SOT are in line with the findings of \citeauthor{Artetxe2018a} (see Figure \ref{fig:sot_sdepuk}). For MC+PCR we observe the greatest performance improvement when the number of removed PCs is around $m=\frac{d}{100}$ (see Figure \ref{fig:moc_sdepuk}). This fits the rule of thumb as stated by \citeauthor{mu2018allbutthetop}.

\subsection{LSCD}

\begin{figure*}[ht]
    \begin{subfigure}{0.33\textwidth}
        \includegraphics[width=\linewidth]{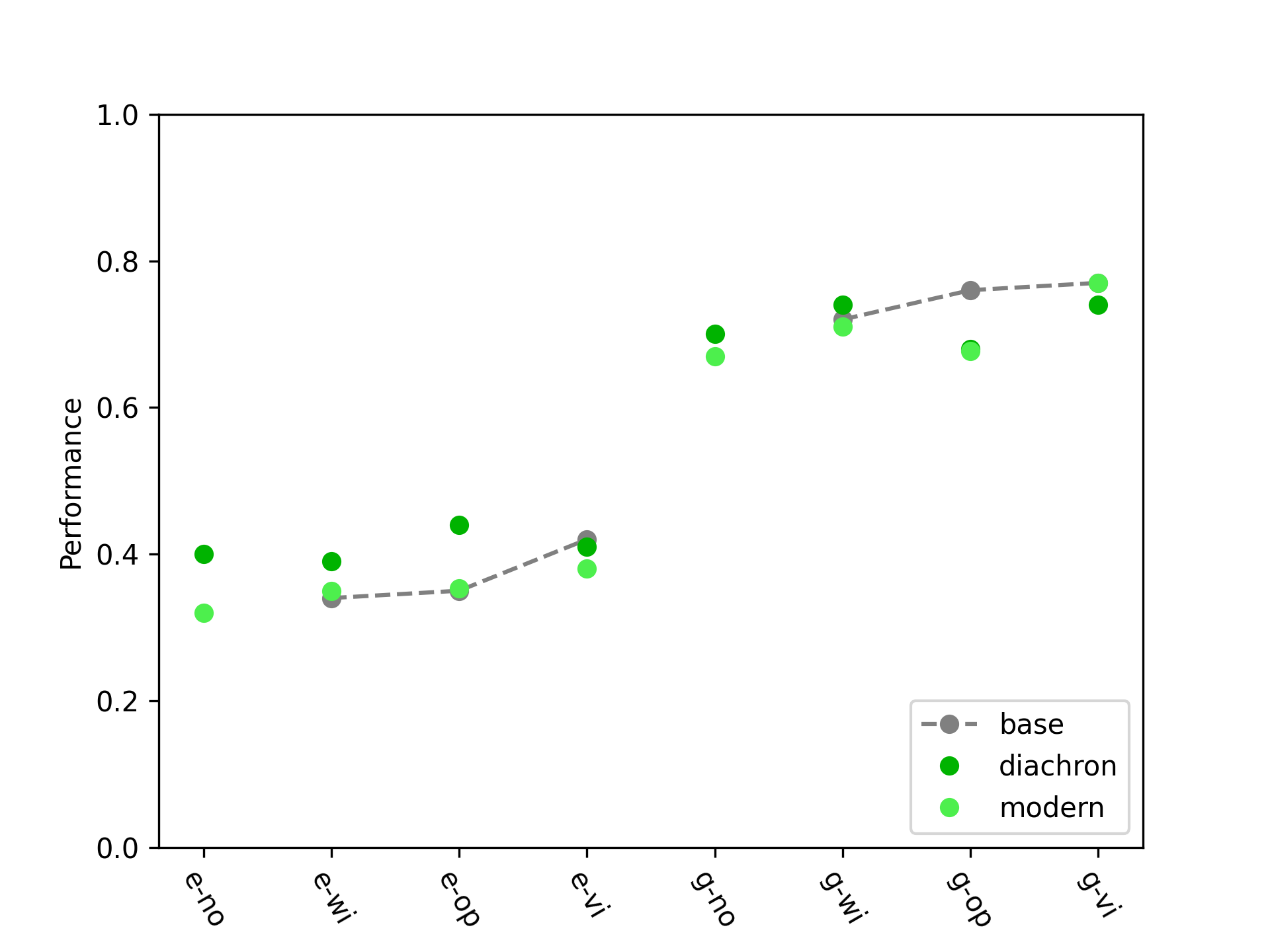}
        \label{fig:LSCDPerformance}
            \end{subfigure}
    \begin{subfigure}{0.33\textwidth}
        \includegraphics[width=\linewidth]{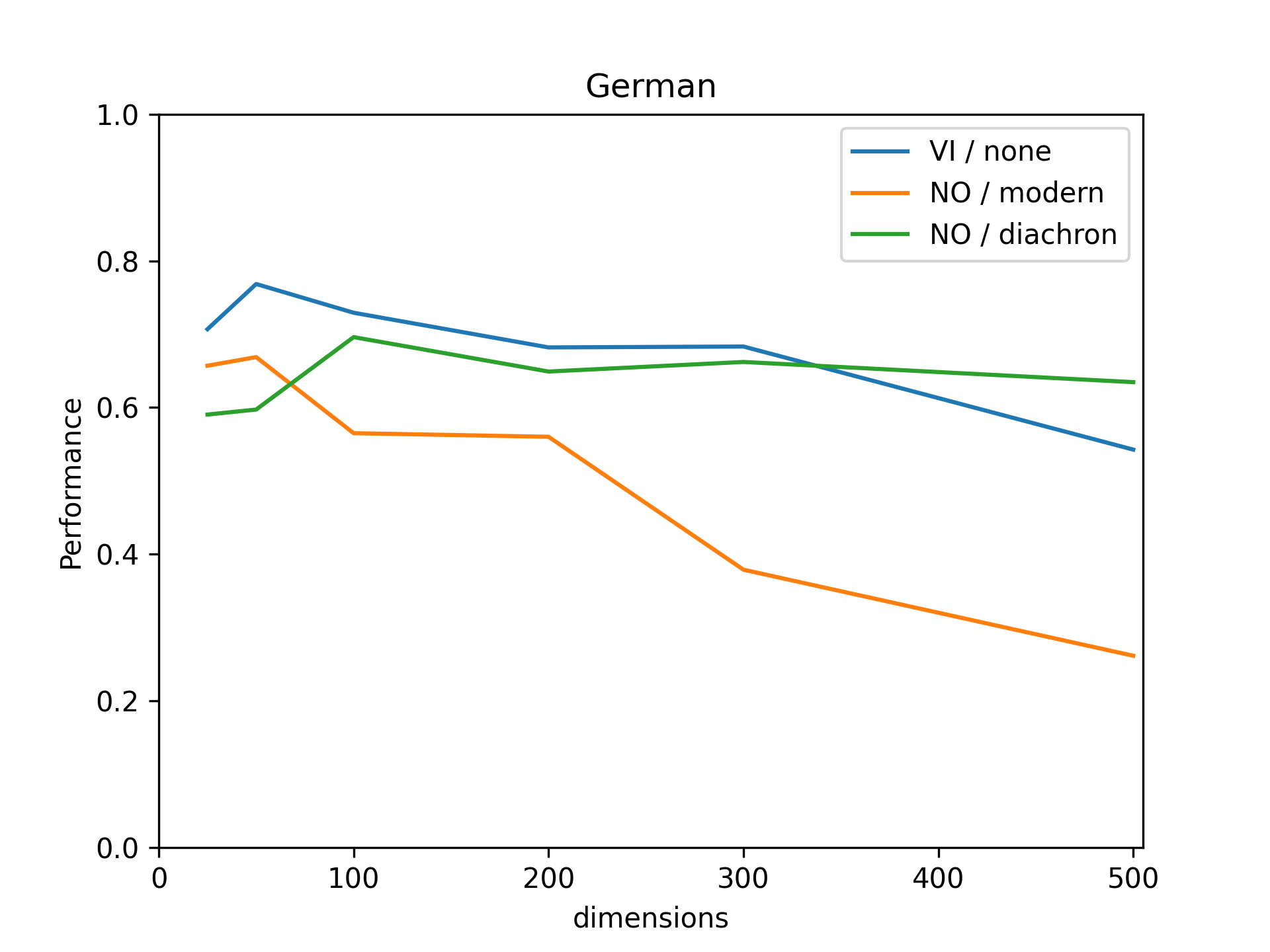}
        \label{fig:pretraincompGERNOVI}
            \end{subfigure}
    \begin{subfigure}{0.33\textwidth}
        \includegraphics[width=\linewidth]{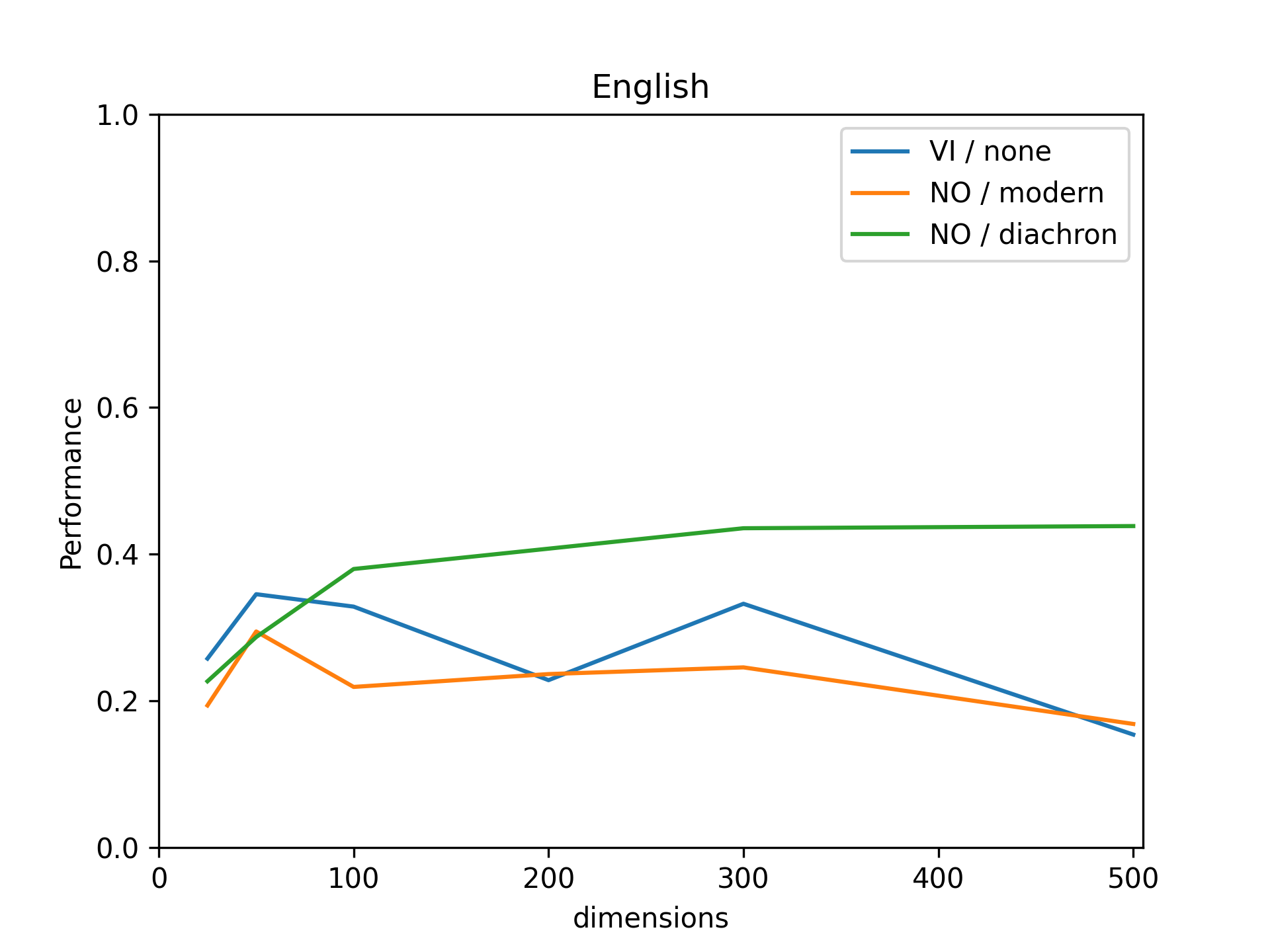}
        \label{fig:pretraincompENGNOVI}
            \end{subfigure}
    \caption{\textbf{Left}: max scores from Table \ref{tab:pretrain}, \textbf{middle} and \textbf{right}: Performance (Spearman's rho) of NO alignment method on LSCD task across different dimensionalities and pre-training corpora. VI without pre-training as comparable baseline.}
    \label{fig:pretaincomp}
\end{figure*}

\subsubsection{Pre-training}
We tune SGNS models for each alignment method with and without pre-training (baseline), see Table \ref{tab:pretrain}. Recall from Section \ref{sec:pretraindescription} that we use the corpora \textsc{modern} and \textsc{diachron} for pre-training.
Table \ref{tab:pretrain} lists the maximum and mean performances of the baseline and pre-training with different alignment methods, as well as the standard deviation (for a visual representation of the max values see Figure \ref{fig:pretaincomp}). The mean is calculated across different $d$, $e$ and $w$, giving the expected performance in a realistic scenario where fine-tuning hyper-parameters is not possible \citep{schlechtweg2020semeval,diacrita_evalita2020}. For German, the baseline max and mean scores could not be significantly improved by pre-training across alignments.
For English, pre-training on \textsc{diachron} results in better max and mean scores for OP and WI, with max improvements up to .10. Also, the overall best result is achieved with OP and pre-training on \textsc{diachron}. The usage of \textsc{modern} does not improve on the maximum, while reducing the mean. The overall lower performance as well as the observed performance improvements compared to German, may be attributed to the roughly 10 times smaller target corpora. That is, pre-training is helpful on the smaller target corpora.

\begin{table*}[t]
\center
\begin{tabular}{ c | c | c c | c c | c c }
\toprule
& \multirow{2}{*}{align.} & \multicolumn{2}{c|}{baseline} & \multicolumn{2}{c|}{\textsc{diachron}} &\multicolumn{2}{c}{ \textsc{modern}} \\
& & \textit{max} & \textit{mean/std} & \textit{max} & \textit{mean/std} & \textit{max} & \textit{mean/std} \\
\hline
\multirow{4}{*}{\rotatebox[origin=c]{90}{\textbf{GER}}} & VI & \textbf{.77} & .72 / .063$^*$ & .74 & .61 / .067$^*$ & \textbf{.77} & .70 / .060$^*$\\
 & OP & \textbf{.72} & .69 / .022 & .68 & .59 / .049$^*$ & .68 & .61 / .051$^*$ \\
 & WI & \textbf{.76} & .70 / .033 & .74 & .69 / .037$^*$ & .71 & .66 / .043$^*$ \\
 & NO & - & - / - & .70 & .58 / .081$^*$ & .67 & .60 / .050$^*$\\
\midrule
\multirow{4}{*}{\rotatebox[origin=c]{90}{\textbf{ENG}}} & VI & \textbf{.42} & .30 / .067 & .41 & .28 / .073 & .38 & .26 / .060\\
 & OP & .34 & .28 / .041 & \textbf{.44} & .31 / .071 & .35 & .27 / .047\\
 & WI & .35 & .28 / .041 & \textbf{.39} & .29 / .053 & .35 & .24 / .055\\
 & NO & - & - / - & .40 & .34 / .080 & .32 & .24 / .060\\
\bottomrule
\end{tabular}
\caption{max and mean performance on LCSD task (Spearman's rho) for all alignment methods. Note: mean values marked with ($^*$) ignore results utilizing $d$ $<$100 due to consistent performance drops at higher $d$.}
\label{tab:pretrain}
\end{table*}

\subsubsection{Post-processing}
\label{sec:post} 

\begin{figure*}[ht]
    \center
    \begin{subfigure}{0.32\textwidth}
        \includegraphics[width=\linewidth]{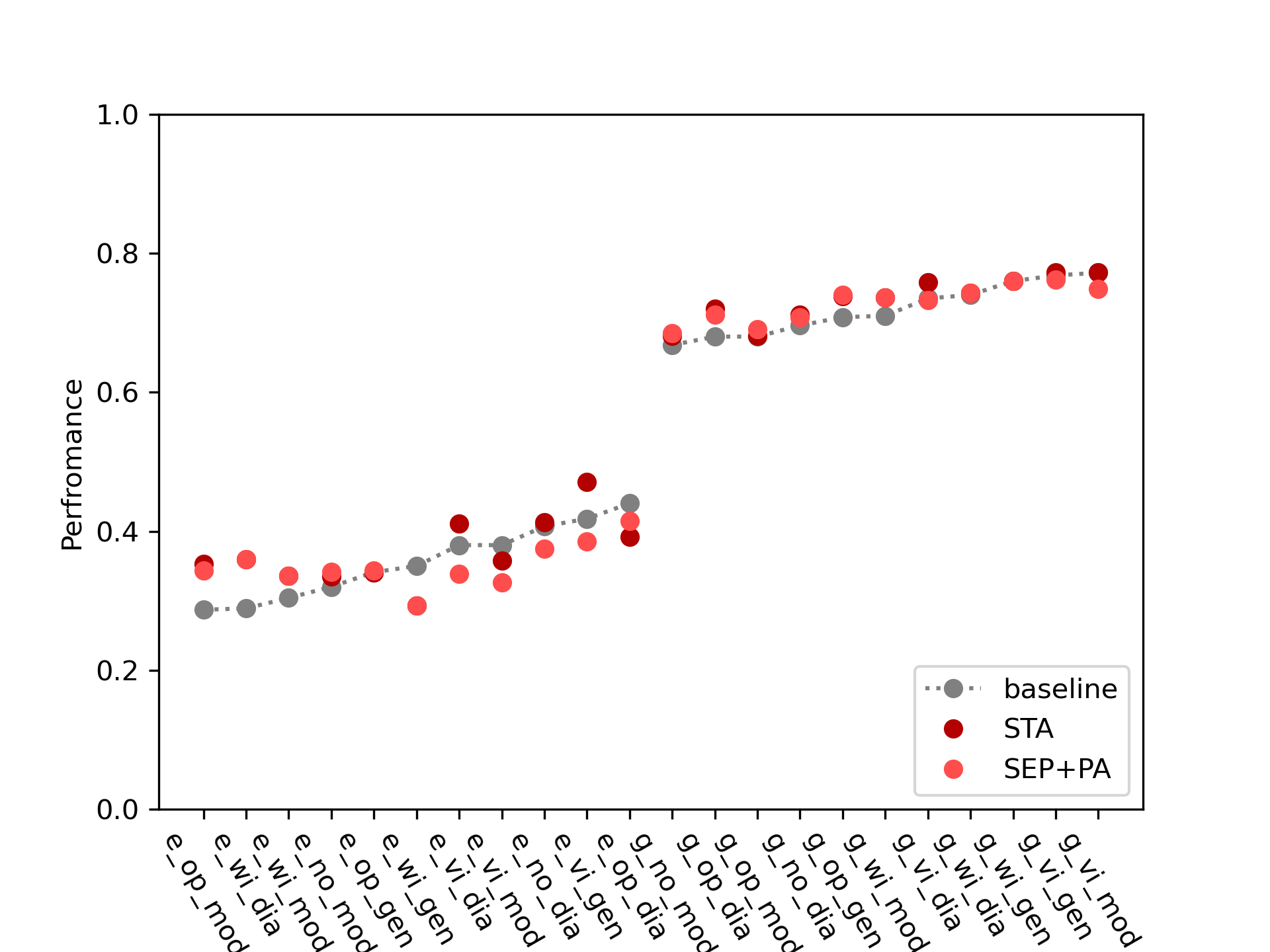}
        \caption{}
        \label{fig:highscore_sto}
    \end{subfigure}
    \begin{subfigure}{0.32\textwidth}
        \includegraphics[width=\linewidth]{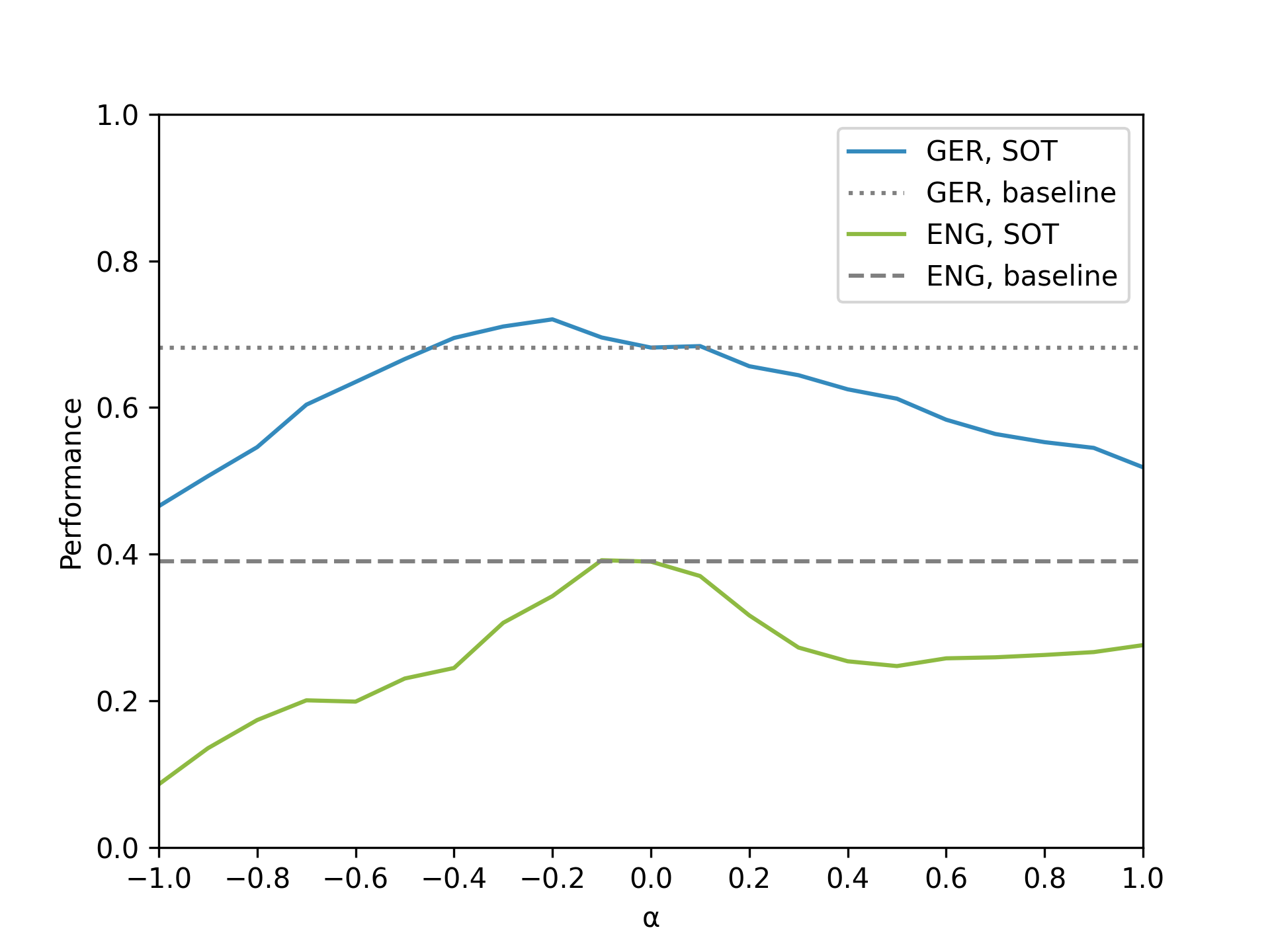}
        \caption{}
        \label{fig:ger_eng_rep}
    \end{subfigure}
    \begin{subfigure}{0.32\textwidth}
        \includegraphics[width=\linewidth]{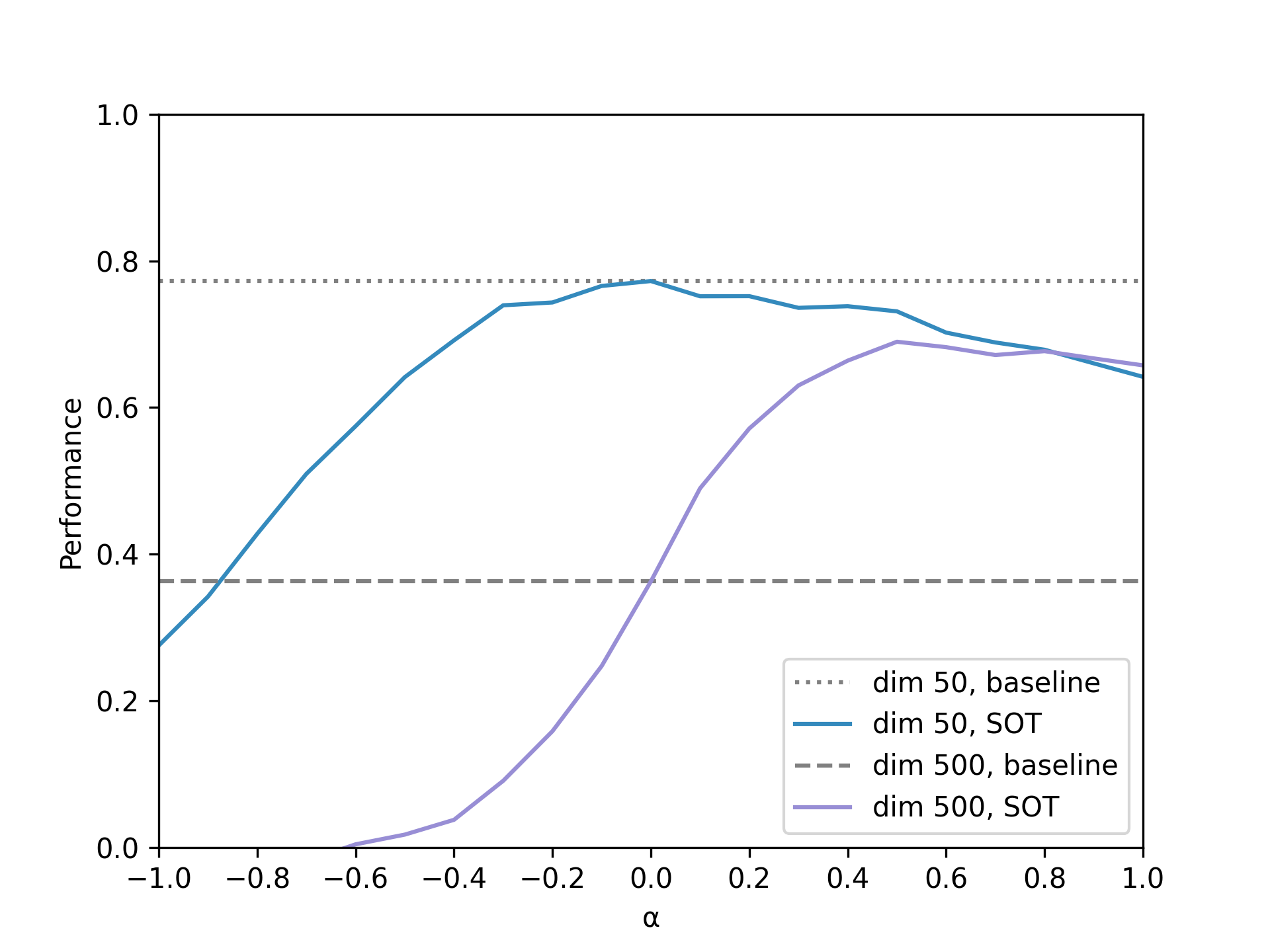}
        \caption{}
        \label{fig:vi-big}
    \end{subfigure} 
    \begin{subfigure}{0.32\textwidth}
        \includegraphics[width=\linewidth]{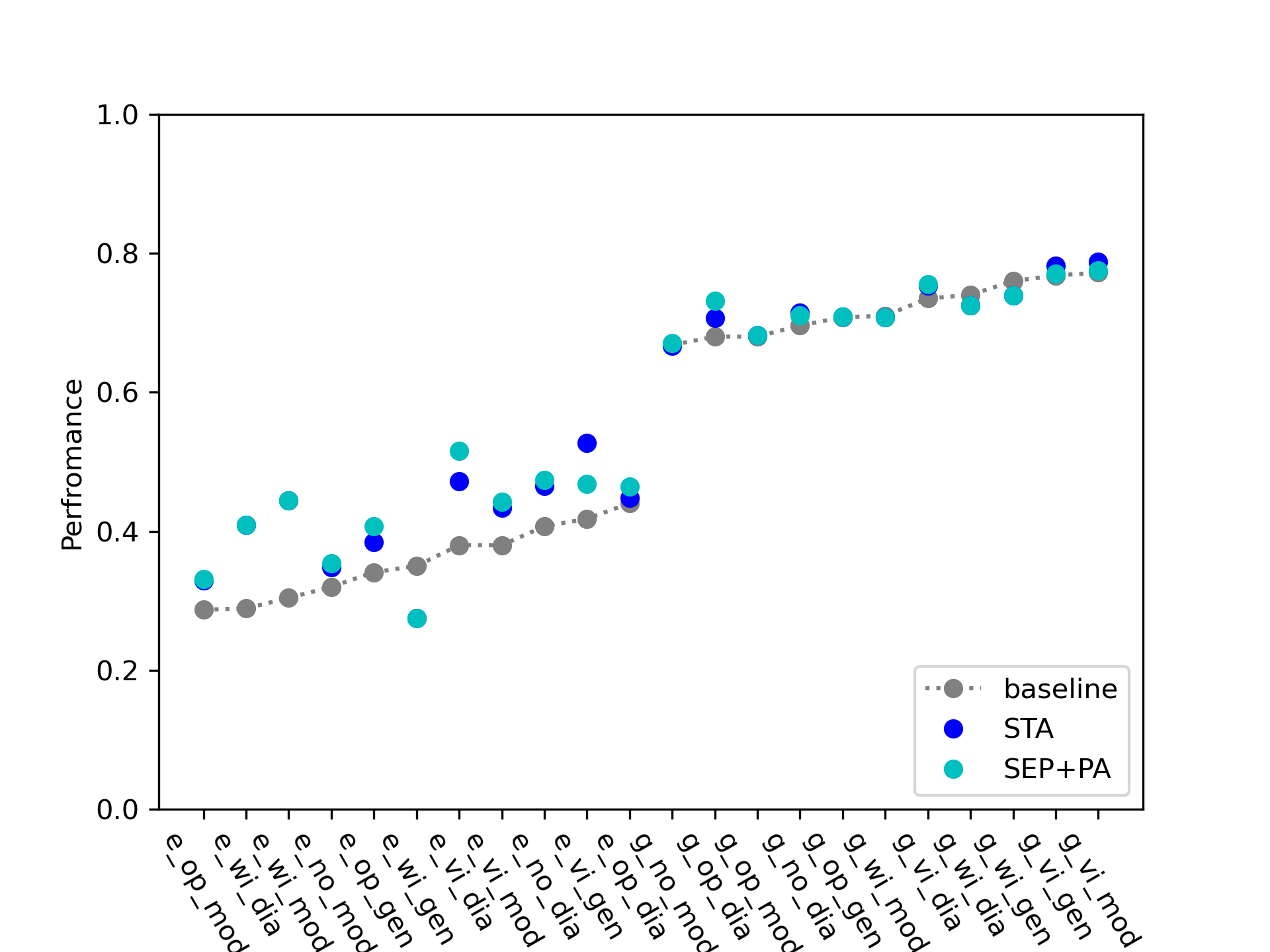}
        \caption{}
        \label{fig:highscore_mcpcr}
    \end{subfigure}
    \begin{subfigure}{0.32\textwidth}
        \includegraphics[width=\linewidth]{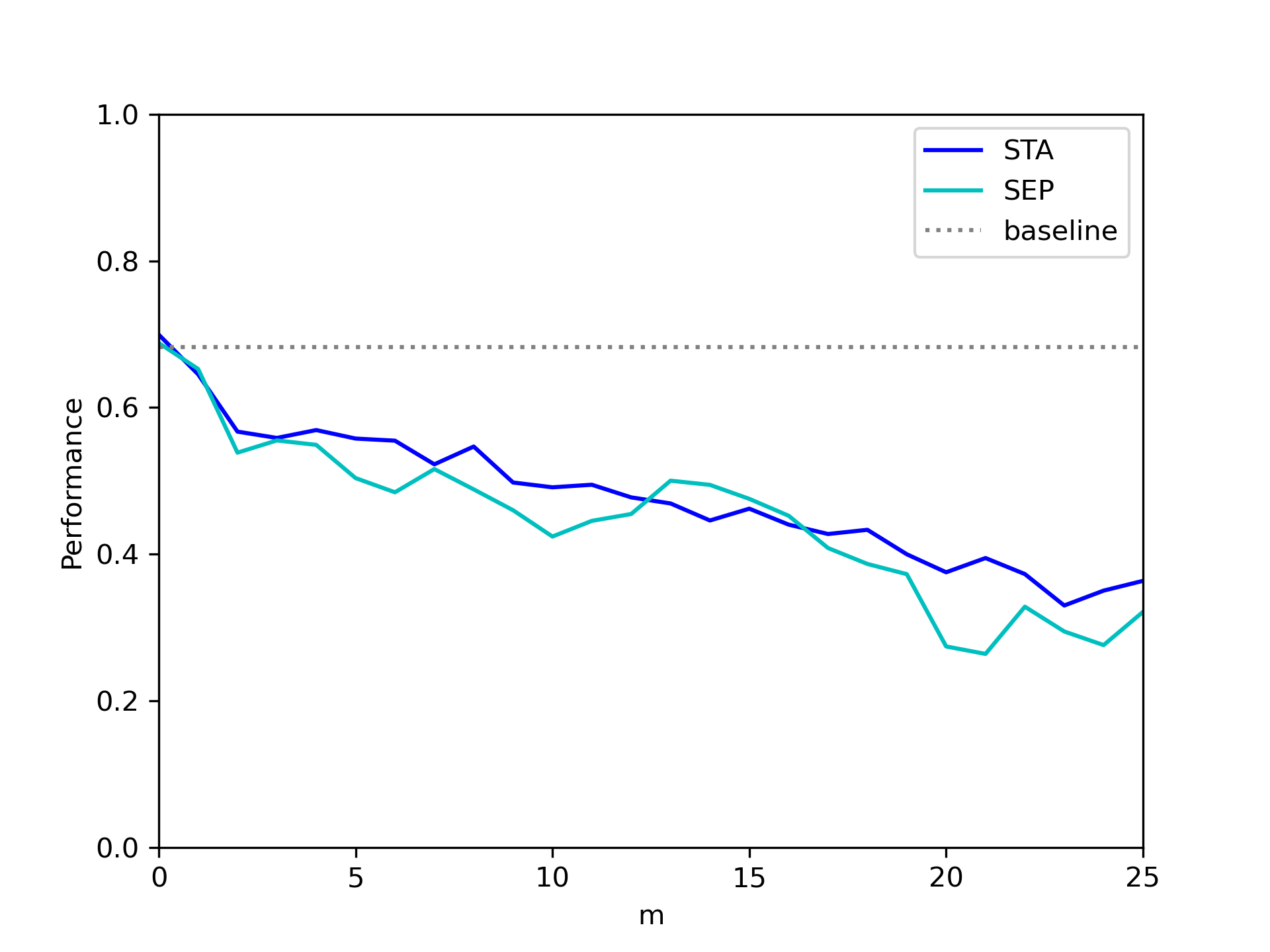}
        \caption{}
        \label{fig:MCPCRgerGen200}
    \end{subfigure}
    \begin{subfigure}{0.32\textwidth}
        \includegraphics[width=\linewidth]{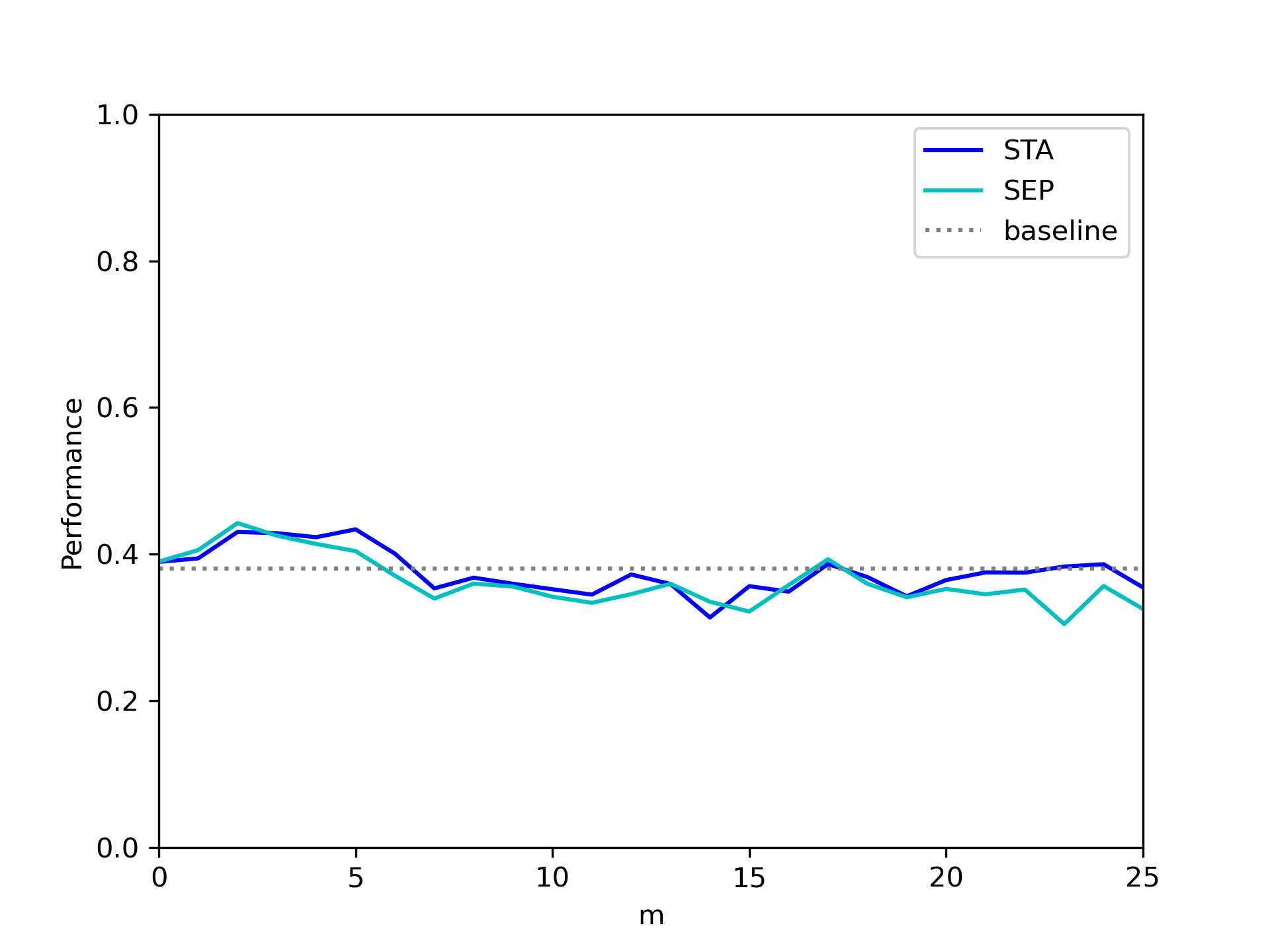}
        \caption{ }
        \label{fig:MCPCRengBig300}
    \end{subfigure} 
    \caption{\textbf{Top:} SOT (\ref{fig:highscore_sto}, \ref{fig:vi-big}, \ref{fig:ger_eng_rep}), \textbf{Bottom:} MC+PCR (\ref{fig:highscore_mcpcr}, \ref{fig:MCPCRgerGen200}, \ref{fig:MCPCRengBig300}). Performance over high-scores (\ref{fig:highscore_sto}, \ref{fig:highscore_mcpcr}).
    Representative results after SOT+STA over german and english dataset (\ref{fig:ger_eng_rep}). Representative plot of ``repair'' effect after SOT+STA for GER+VI (\ref{fig:vi-big}).
    Representative result after MC+PCR over German and English dataset (\ref{fig:MCPCRgerGen200}, \ref{fig:MCPCRengBig300}). Note for \ref{fig:highscore_sto} and \ref{fig:highscore_mcpcr}: where data points overlap only lighter colour visible; dashed line between baseline data points only a visual aid.}
    \label{fig:repdata}
\end{figure*}

For every combination of alignment and pre-training method, the matrix with the highest performance across parameters is chosen as the baseline. SOT and MC+PCR are applied individually to these matrices within a wide parameter range (see Appendix \ref{sec:parameters}) for both stacking methods (STA and SEP/SEP+PA). Table \ref{tab:perfargmax} presents the mean optimal performance gains after PP, which is calculated by extracting the best performance after PP for every matrix, subtracting the baseline values and averaging the values per language. Averaging the respective parameter values yields the mean argmax. Figure \ref{fig:highscore_sto} and \ref{fig:highscore_mcpcr} show the highest performances for every baseline matrix after SOT and MC+PCR respectively. 
\paragraph{SOT.} 
As we see in Figure \ref{fig:highscore_sto}, SEP+PA and STA perform similarly. We find small mean performance gains across the board (.013 for GER+STA, .008 for GER+SEP+PA, .013 for ENG+STA), except for ENG+SEP+PA where a minuscule decrease (-.005) can be seen. Overall, STA outperforms SEP+PA slightly. We now further examine the effect of SOT+STA on individual matrices. In general, the data can approximately be described as a downward opening parabola (see Figure \ref{fig:ger_eng_rep}), with different peaks for both languages and slight differences between alignment methods. Averaging the argmax for $\alpha$ shows us where these peaks are. The calculations yield a mean optimal $\alpha$ of 0 for GER+STA, and -0.2 for ENG+STA. For GER the peak performance always lies in the interval $[-0.2, 0.3]$. This changes to $[-0.4,0.1]$ for ENG, except for one outlier, where the peak is at -0.8. Moving $\alpha$ away from this parameter range results in severe performance decreases. This behaviour can also be seen on the \textsc{modern} corpora (see Figure \ref{fig:sde_puk}) and is in line with the findings of \citet{artetxe-etal-2018-uncovering}. In order to predict a high-performing parameter, independent from the underlying matrix, we calculate mean performance gains for fixed parameter values. The values are chosen according to the the above-described peak intervals for the respective languages. However, on average, using a fixed parameter results in slight performance losses, notwithstanding the $\alpha$-value, and hence, finding a high-performing fixed parameter value was not possible. We observe similar findings for individual alignment methods and varying dimensionality. However, GER+VI alignment represents an interesting exception: With high dimensionality ($>$ 300) base performance drops heavily \citep{kaiser-etal-2020-IMS}, and is then ``repaired'' by the PP, bringing it close to the baseline of the best performing dimension (see Figure \ref{fig:vi-big}).

\begin{table}[t]
\center
\small
\begin{tabular}{ c | l | r r }
\toprule
    & \multirow{2}{*}{PP + STA/SEP}   & argmax & gain \\
    &  & \textit{mean/std} & \textit{mean/std} \\
    \hline
    \multirow{4}{*}{\rotatebox[origin=c]{90}{\textbf{GER}}} 
    & SOT + STA     & 0.0/0.2   & .013/.013  \\
    & SOT + SEP+PA  & 0.1/0.3   & .008/.015  \\
    & MC+PCR + STA  & 1.2/1.6   & .004/.042  \\
    & MC+PCR + SEP  & 0.7/1.1   & .004/.043  \\
    \midrule
    \multirow{4}{*}{\rotatebox[origin=c]{90}{\textbf{ENG}}} 
    & SOT + STA     & -0.2/0.2  & .013/.041  \\
    & SOT + SEP+PA  & -0.2/0.3  & -.005/.043 \\
    & MC+PCR + STA  & 3.0/3.8  & .049/.068  \\
    & MC+PCR + SEP  & 6.2/7.1  & .058/.077  \\
\bottomrule
\end{tabular}
\caption{Mean of best-performing parameters and mean performance gain compared to baseline on LSCD task. Parameter range for SOT [-1,1] and MC+PCR [0,25]}
\label{tab:perfargmax}
\end{table}

\paragraph{MC+PCR.} 
As we see in Figure \ref{fig:highscore_mcpcr}, MC+PCR yields small improvements over the baselines for German. This is also reflected in the mean gain in Table \ref{tab:perfargmax}. 
We find that no single value for $m$ yields consistent improvements. However, we find that for $m$=0 (only MC) MC+PCR consistently improves the baseline slightly (see Figure \ref{fig:MCPCRgerGen200}), while for higher $m$ the performance decreases consistently.
For English we see greater improvements, see Figure \ref{fig:MCPCRengBig300}, \ref{fig:highscore_mcpcr} and mean gain in Table \ref{tab:perfargmax}. A range of parameters shows improvements with $m$=3 yielding the highest (.0175). This can also be seen in Figure \ref{fig:MCPCRengBig300} where several parameters yield improvements.
We conclude that predicting a parameter for likely performance improvement is possible for English, but not for German. However, if this PP should be used, we recommend using a parameter space of $m\in$ [0, 5], as this parameter space is most likely to produce improvements on English, while not harming performance too much on German. This also roughly corresponds to the recommendation of \citet{mu2018allbutthetop}, as they predict that the parameter should be chosen around $\frac{d}{100}$. Furthermore, we suggest using STA, as this does on average show better performance over SEP for the aforementioned parameter space. We see that the effects of SOT as well as MC+PCR are highly dependent on the underlying matrix.

\section{Analysis}
\label{sec:analysis}
\paragraph{Test Statistics.}
The effects of pre-training and PP methods on word embeddings are not limited to performance differences in word similarity or LSCD tasks. We use two test statistics to further analyse vector spaces: (i) isotropy \cite{mu2018allbutthetop}, i.e., uniformity of vector distribution and (ii) frequency bias \cite{dubossarsky2017,kaiser-etal-2020-IMS}, i.e., correlation between cosine distance and frequency.\footnote{We compute correlation based on frequency in the second target corpus, results were similar for the first target corpus.}

\subsection{Pre-training}
\begin{figure*}[ht]
    \begin{subfigure}{0.33\textwidth}
        \includegraphics[width=\linewidth]{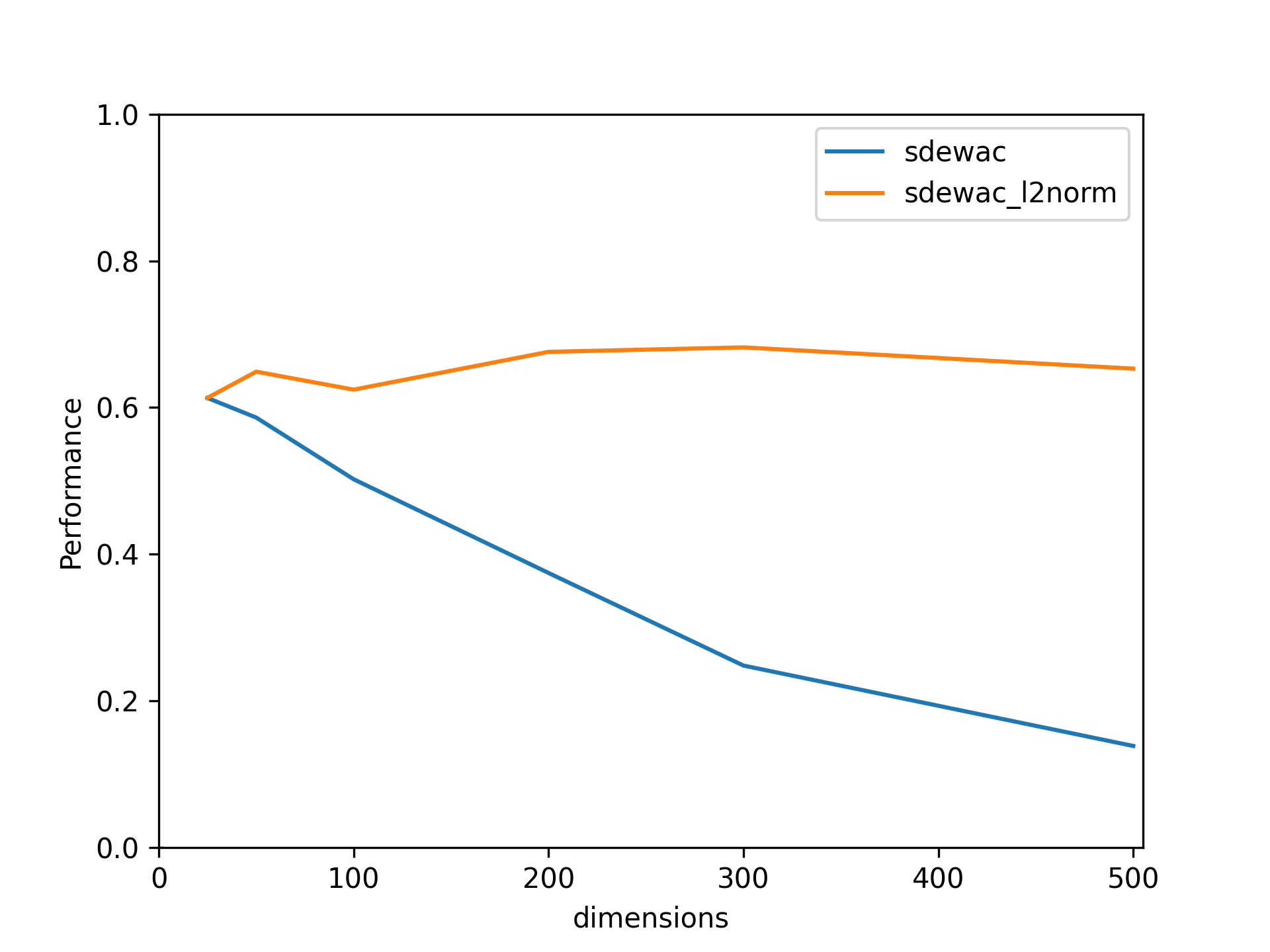}
        \caption{}
        \label{fig:sde_fsl2_performance_lowfreq}
    \end{subfigure}
    \begin{subfigure}{0.33\textwidth}
        \includegraphics[width=\linewidth]{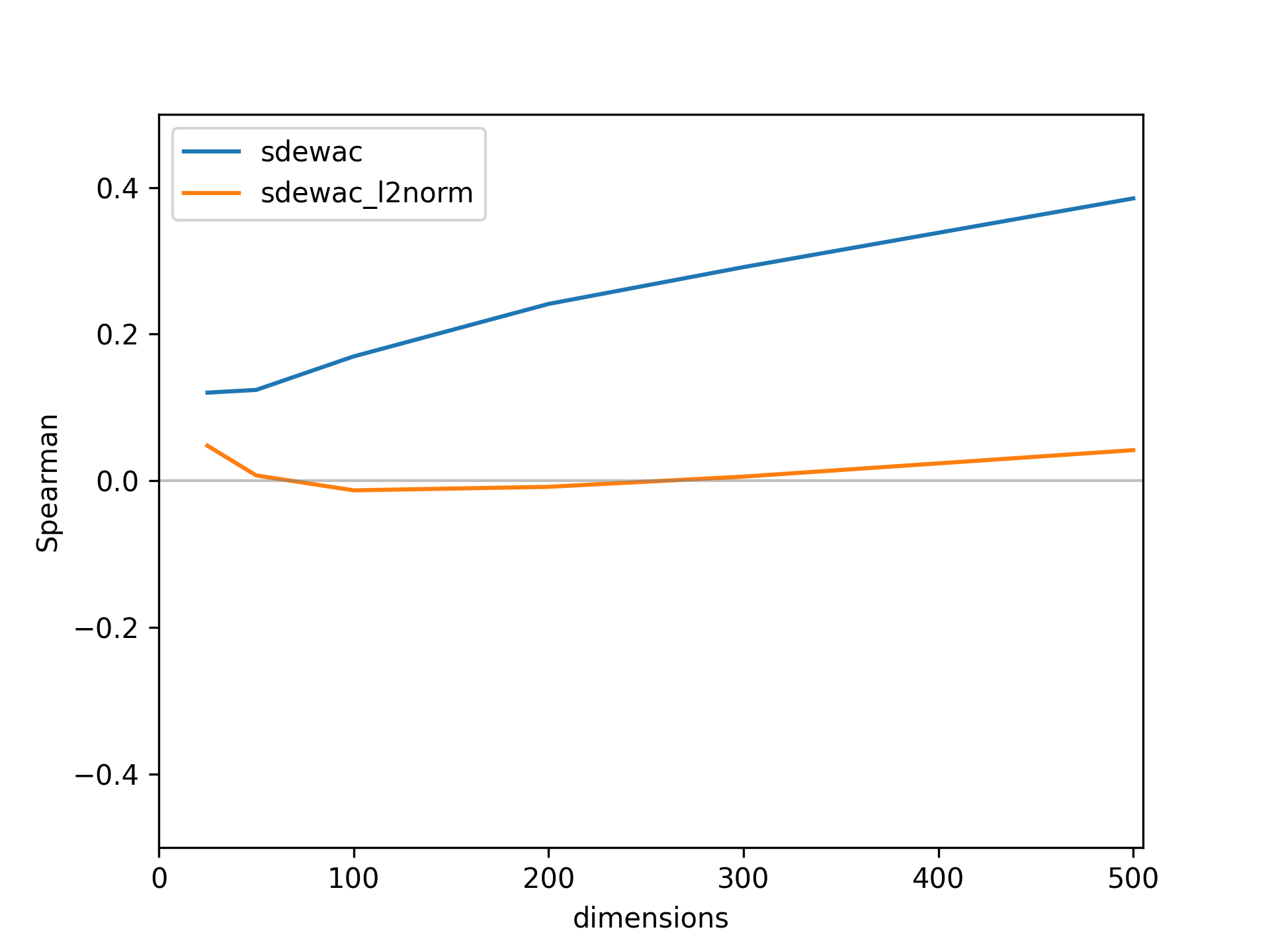}
        \caption{}
       \label{fig:sde_freqcorr}
    \end{subfigure}
    \begin{subfigure}{0.33\textwidth}
        \includegraphics[width=\linewidth]{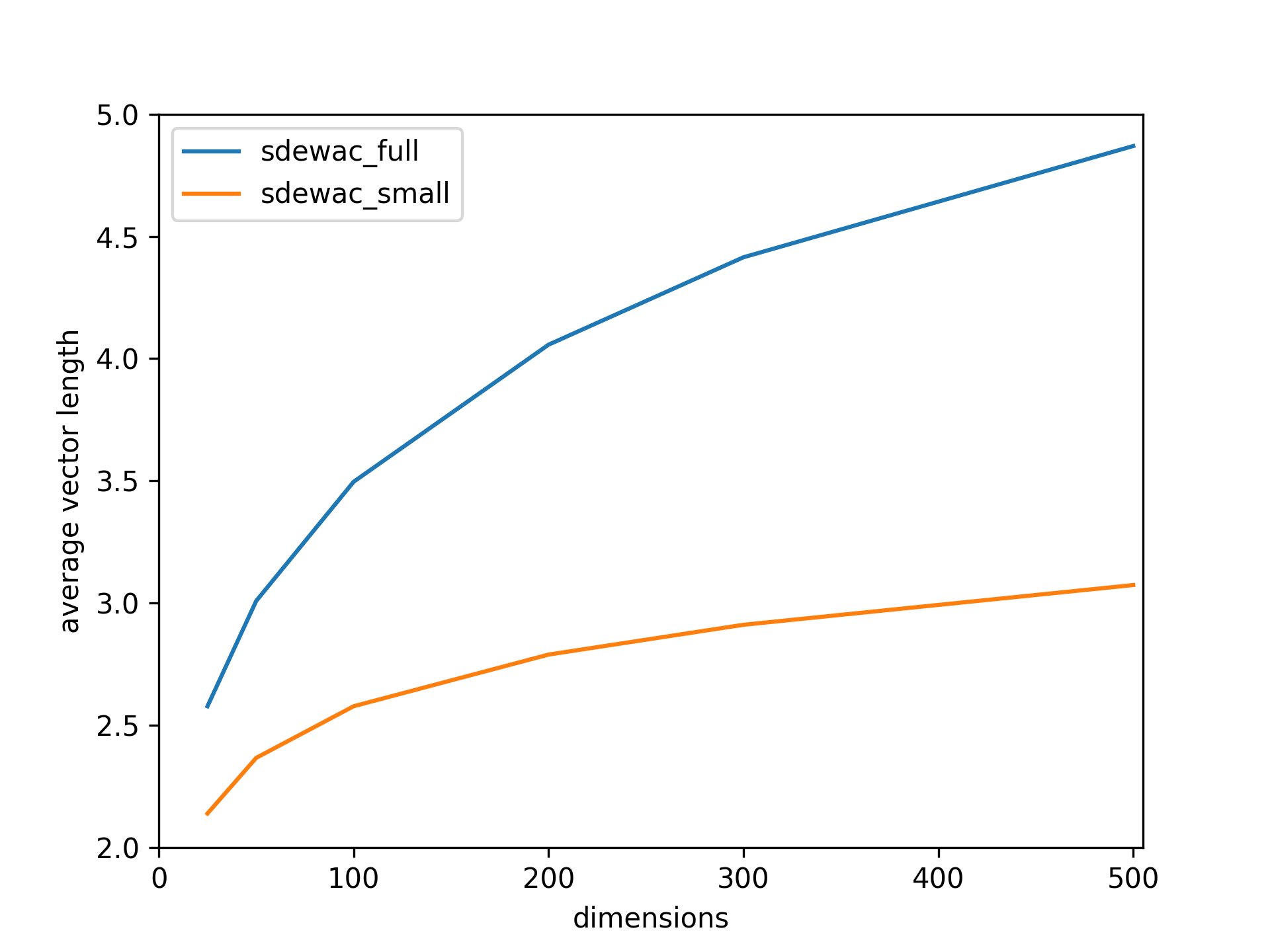}
        \caption{}
        \label{fig:sde_fl2_vec_length}
    \end{subfigure}
    \caption{Test-statistics for result analysis, \textbf{left}: Performance after pre-training on different corpora, \textbf{middle}: Correlation between CD and frequency, \textbf{right}: average vector length of the weights created on different-sized pre-training corpora.}
    \label{fig:pretrainanalysis}
\end{figure*}

\label{sec:pretraininganalysis}
On the German dataset it is noticeable that pre-training on \textsc{diachron} often results in slight drop in performance at higher $d$. This behaviour is more pronounced, consistent and even visible on the English dataset when pre-training on \textsc{modern}, see Figure \ref{fig:pretaincomp}.\footnote{Although not depicted, the other alignment techniques in combination with pre-training show very similar behaviour to NO.} 
Such a drop in performance after initializing on pre-trained vectors has already been observed by \citet{kaiser-etal-2020-IMS}. The authors relate the drop to an increased frequency bias and reduce it by increasing $e$/$w$. It is noteworthy that the drop is much more pronounced for pre-training on \textsc{modern} compared to \textsc{diachron}. This can be attributed to a difference in word vector lengths of the SGNS model used for initialization. We make the following observation: average word vector length increases with the amount of training word pairs. The difference more training data makes is amplified at higher $d$, see Figure \ref{fig:sde_fl2_vec_length}. By length-normalizing the word vectors between the initialization and training step, the drop in performance can be completely circumvented. Additionally, the frequency bias is reduced to 0, see Figure \ref{fig:sde_freqcorr}.

For English, we expected a higher performance gain from pre-training when using \textsc{modern} because of the small data size. However, we observe no improvements over the baseline. Using length-normalized word vectors for initialization does result in slightly improved max and mean values for \textsc{modern} but these are still lower than max and mean values of \textsc{diachron}.

\subsection{SOT}

\begin{figure*}[ht]
\centering
    \begin{subfigure}{0.4\textwidth}
        \includegraphics[width=\linewidth]{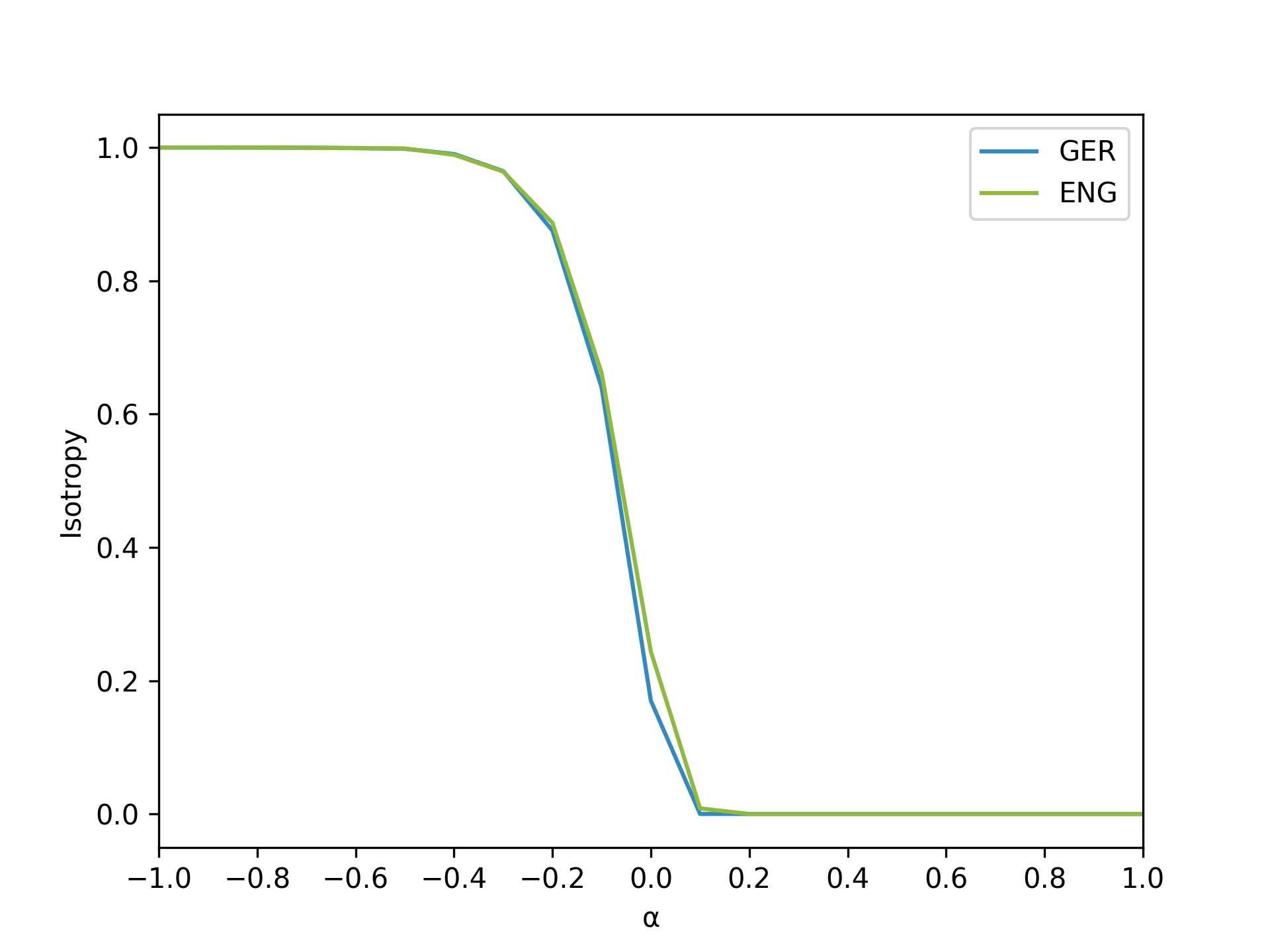}
        \caption{}
        \label{fig:plotiso}
    \end{subfigure}
    \hspace{2cm}
    \begin{subfigure}{0.4\textwidth}
        \includegraphics[width=\linewidth]{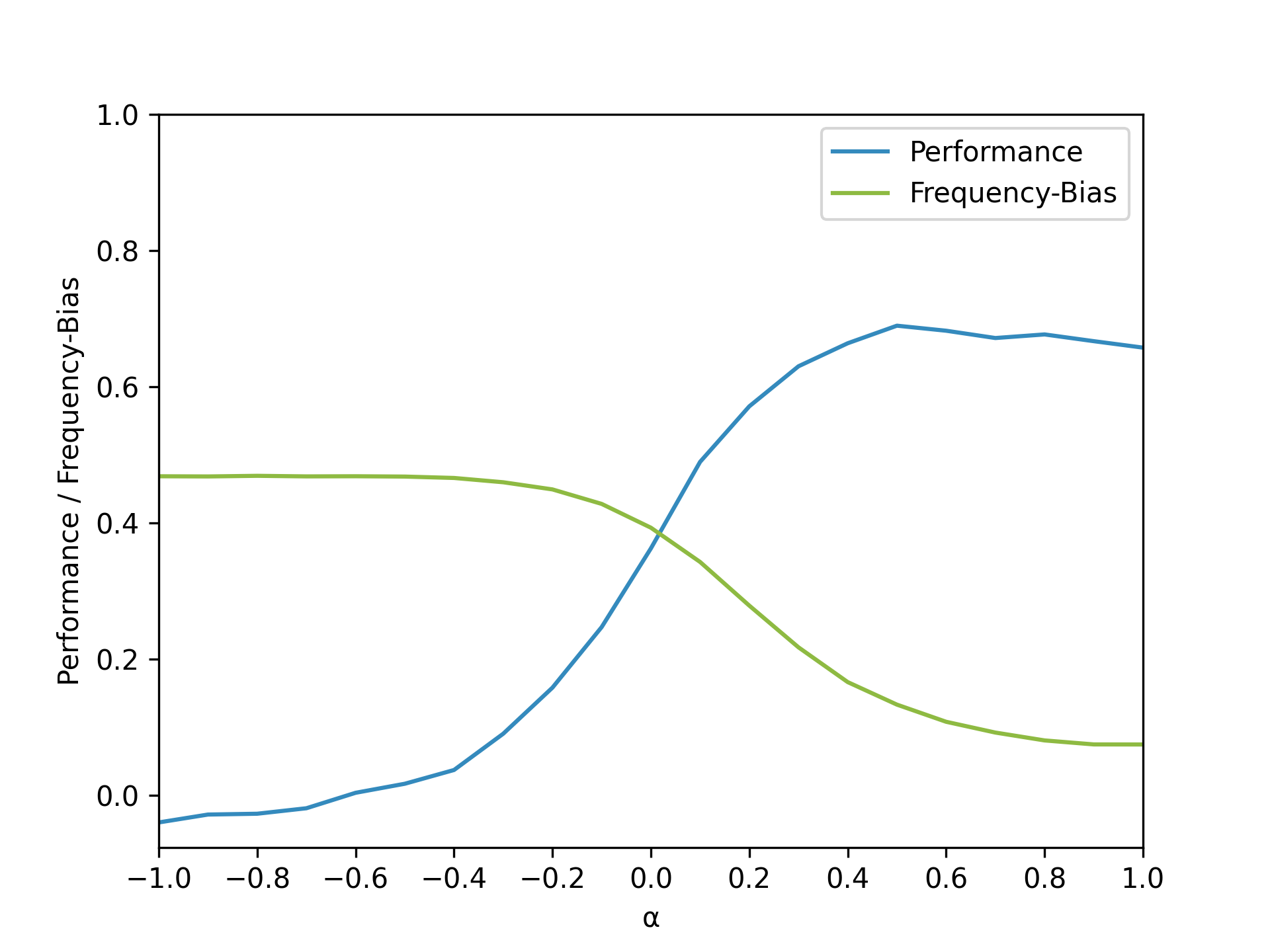}
        \caption{}
        \label{fig:perf_freq}
    \end{subfigure}
    \caption{Representative plot for the isotropy after SOT+STA (\ref{fig:plotiso}). Performance and frequency bias after SOT+STA for GER+VI+BIG (\ref{fig:perf_freq}).}
    \label{fig:label2}
\end{figure*}

SOT has a clear effect on isotropy, which has not been described in previous research. Isotropy shows the same behaviour across both languages and all models, and is best described as a vertically mirrored S-curve (see Figure \ref{fig:plotiso}). Decreasing $\alpha$ increases isotropy close to 1, while increasing $\alpha$ decreases isotropy close to 0. The average correlation (Pearson) between $\alpha$ and isotropy over all matrices is -.89 for both languages. However, the performance correlates only slightly with isotropy (-.25, .35). Moreover, $\alpha$ correlates only weakly with frequency bias (.19, -.12, however with high variance). In order to explain the above-described ``repair'' effect we take a closer look at the three GER+VI models. Applying SOT brings large performance increases, as stated in Section \ref{sec:post}. For all three models a considerably higher baseline frequency bias for $d$=500 is visible. SOT strongly reduces this bias for \textsc{modern}, and results in a huge performance gain (see Figure \ref{fig:perf_freq}). 

\subsection{MC+PCR}
As \citet{mu2018allbutthetop}'s main motivation behind MC+PCR is to increase isotropy of a vector space as well as removal of word frequency noise through PCR, we examine how isotropy and frequency bias develop with $m$. While PCR has the predicted effect on frequency bias (GER: -.94, ENG: -0.6), PCR does in fact not increase isotropy, contrary to \citeauthor{mu2018allbutthetop}'s motivation of ``rounding towards isotropy", but has a consistent reducing effect (GER -.75, ENG: -.7). Thus, we believe that rounding towards isotropy is not suitable for explaining performance. Furthermore, we observe that MC not only exhibits effects on isotropy, but also acts on frequency bias, thus \citeauthor{mu2018allbutthetop}'s PCR motivation can be extended to MC.

\section{Conclusion}
We tested the effects of pre-training and post-processing on a variety of LSCD models. We performed extensive experiments on a German and an English LSCD dataset. 
According to our findings, pre-training is advisable when the target corpora are small and should be done using diachronic data. The size of the pre-training corpus is crucial, as a large number of training pairs leads to performance drops, which are probably caused by their effect on vector length. Length-normalization may be used on pre-trained vectors to counteract this effect. 

Further performance improvements may be reached by post-processing. While SOT+STA yielded moderate improvements for both languages, MC+PCR showed larger improvements, but only on English. However, for neither we were able to find a reliable parameter that performed well across the board. Instead, we found that a well-performing parameter value is highly dependent on the underlying matrix. Both post-processing methods affect isotropy and frequency bias. 

The methods we tested are particularly helpful when tuning data is available, as performance can be optimized and becomes more predictable. Hence, we recommend to obtain a small annotated sample of target words for the target corpora and to tune pre-training, model and post-processing parameters on the sample before performing predictions for semantic changes on unseen data. With the recent upsurge of digitized historical corpora and diachronic semantic annotation efforts \citep{Tahmasebi17,Schlechtwegetal18, schlechtweg2020semeval, diacrita_evalita2020, rodina2020rusemshift} this may often be a likely and feasible scenario.

\section*{Acknowledgments}
Dominik Schlechtweg was supported by the Konrad Adenauer Foundation and the CRETA center funded by the German Ministry for Education and Research (BMBF) during the conduct of this study. We thank the reviewers for their insightful feedback.

\bibliography{anthology,biblio}
\bibliographystyle{acl_natbib}

\newpage
\quad
\newpage
\appendix

\section{Corpus details}
The corpora are lemmatized and contain no punctuation, further pre-processing on the corpora by us is limited to removing low-frequency words. All words with a frequency below the value listed in row \textit{min word freq.} in Table \ref{tab:corpora} are removed from the corpora. This is done to reduce noise and unwanted artifacts.

\section{Parameter settings}
\label{sec:parameters}
\paragraph{SGNS.} We use common hyper-parameter settings: initial learning rate of 0.025, number of negative samples $k$=5 and no sub-sampling. Vector dimensionality $d$, window size $w$ and number of training epochs $e$ are varied in order to fine-tune model and methods. This is important as alignment methods like VI are highly dependent on the choice of $e$ and $d$ \citep{kaiser-etal-2020-IMS}. The following values are used: $w \in \{$5, 10$\}$, $e \in \{$5, 10, 20, 30$\}$, $d \in \{$25, 50, 100, 200, 300, 500$\}$. Due to the immense amount of possible parameter combinations we only ran each setting once.

PP was performed on the high-scores of each language, where we differentiate between different combinations of alignment, pre-training as well as if the matrices were STA or SEP post-processed.

\paragraph{SOT.}
As stated in Section \ref{par_stacking}, SEP is used in combination with post-alignment. We apply SOT with $\alpha$ values ranging from -1 to 1 in 0.1 increments on every baseline matrix with $d \in \{$25, 50, 100, 200, 300, 500$\}$. 
\paragraph{MC+PCR.}
MC+PCR is performed using a parameter space of [0, 25] in order to examine the performance development over a growing number of PCs removed. It is important to note that using the parameter 0 results in only applying MC.

\end{document}